\documentclass{vgtc}                          % final (conference style)
\ifpdf%                                % if we use pdflatex
  \pdfoutput=1\relax                   % create PDFs from pdfLaTeX
  \usepackage{graphicx}                % allow us to embed graphics files
  \DeclareGraphicsExtensions{.pdf,.png,.jpg,.jpeg} % for pdflatex we expect .pdf, .png, or .jpg files
\else%                                 % else we use pure latex
  \usepackage{graphicx}                % allow us to embed graphics files
  \DeclareGraphicsExtensions{.eps}     % for pure latex we expect eps files
\fi%

\graphicspath{{figures/}{pictures/}{images/}{./}} % where to search for the images

\usepackage{microtype}                 % use micro-typography (slightly more compact, better to read)
\PassOptionsToPackage{warn}{textcomp}  % to address font issues with \textrightarrow
\usepackage{textcomp}                  % use better special symbols
\usepackage{mathptmx}                  % use matching math font
\usepackage{times}                     % we use Times as the main font
\usepackage{cite}                      % needed to automatically sort the references
\usepackage{tabu}                      % only used for the table example
\usepackage{booktabs}                  % only used for the table example
\usepackage{enumitem}
\usepackage{caption}
\usepackage{subcaption}
\onlineid{0}

\vgtccategory{Research}

\title{The Object at Hand: Automated Editing for Mixed Reality Video Guidance from Hand-Object Interactions}

\author{Yao Lu\thanks{e-mail: yl1220@bristol.ac.uk}\\ %
        \scriptsize University of Bristol %
\and  Walterio W. Mayol-Cuevas\thanks{e-mail: wmayol@cs.bris.ac.uk}\\ \scriptsize University of Bristol% %
    }

\abstract{In this paper, we concern with the problem of how to automatically extract the steps that compose real-life hand activities. This is a key competence towards processing, monitoring and providing video guidance in Mixed Reality systems. We use egocentric vision to observe hand-object interactions in real-world tasks and automatically decompose a video into its constituent steps. Our approach combines hand-object interaction (HOI) detection, object similarity measurement and a finite state machine (FSM) representation to automatically edit videos into steps. We use a combination of Convolutional Neural Networks (CNNs) and the FSM to discover, edit cuts and merge segments while observing real hand activities. We evaluate quantitatively and qualitatively our algorithm on two datasets: the GTEA\cite{li2015delving}, and a new dataset we introduce for Chinese Tea making. Results show our method is able to segment hand-object interaction videos into key step segments with high levels of precision.   

} % end of abstract

\CCScatlist{
  \CCScatTwelve{Human-centered computing}{Human computer interaction (HCI)}{Interaction paradigms}{};
  \CCScatTwelve{Applied computing}{Education}{Computer-assisted instruction}{}
}

\begin{document}

%% The ``\maketitle'' command must be the first command after the
%% ``\begin{document}'' command. It prepares and prints the title block.

%% the only exception to this rule is the \firstsection command
% \firstsection{Introduction}

\maketitle

%% \section{Introduction} %for journal use above \firstsection{..} instead
\section{Introduction}

Guidance is one of the core target applications for Augmented Reality (AR), specifically in industry or to support daily living. AR systems have often been proposed as enablers to perform maintenance and repair tasks and several systems have been devised for this task \cite{platonov2006mobile,hanson2017augmented,COGNITO2015}.

However, content authoring for real-world tasks is one of the principal challenges in the extended Mixed Reality (MR) field, and one that limits the adoption of MR technology in general \cite{zubizarreta2019framework,leelasawassuk2017automated}. Overall, automating content authoring for MR systems has received relatively little attention in contrast to the body of work that explores how such systems could be used. Commonly, content creation assumes pre-definition of object and scene model assets. This makes content authoring expensive, inefficient and restricted to the manual workflow and content authoring tools available. It is also common to impose strict sequences of pre-defined steps on activities. However, it is precisely real-world tasks such as those in daily living, that have limited pre-existing assets and are largely unscripted that can be helped the most by systems that support memory loss, DIY tutoring, maintenance and assembling tasks. In this paper we argue that it is thus better to work-backwards and observe the world as-is, extracting the task steps and relevant objects that real people use in real tasks and in the wild.

On the other hand, First Person View (FPV) content is produced with increased ease due to the fast development of wearable devices and video-based social media. An egocentric system follows the user's viewpoint, helping to record or deliver data for tutoring of a task to the same or different user. Overall, humans are well-tuned to imitate from watching and learning from imitation, a survival competence that is present in other animals too \cite{kis2015social}. In terms of recent video guidance systems for MR, the system in \cite{leelasawassuk2017automated} showcased that using egocentric video guidance is easier to produce and more accessible for users to follow instructions while freeing their hands. In \cite{lu2019higs}, results show a video-based AR system able to show guidance and monitor hands but from pre-defined hand poses and locations. In this work, we expand from prior MR content generation work by focusing on the critical aspect of automated task decomposition via hand-object interaction detection.

%Importantly, what user receives from the guidance video largely relies on the quality of recordings. In practice, people may perform a same task with different hands, pace, and order. This brings significant uncertainties in guidance delivering. We found the starts and ends of actions are normally accompanied by picking up and putting down of an object. Thus the most important cue in FPV content, we argue, is hand and it's interaction with objects.
\begin{figure}[tb]
 \centering % avoid the use of \begin{center}...\end{center} and use \centering instead (more compact)
 \includegraphics[width=1\columnwidth]{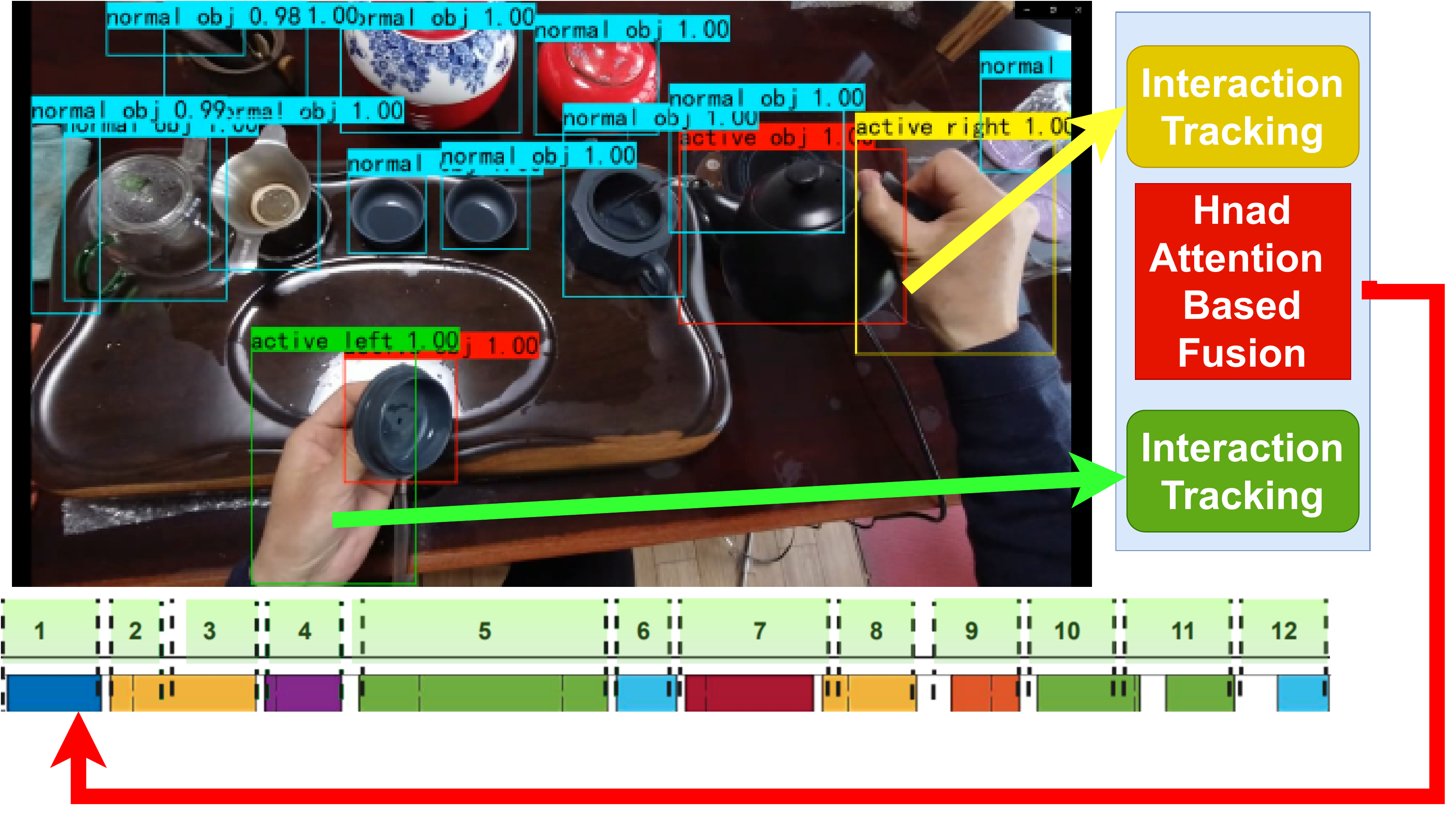}
 \caption{Our hand-object interaction (HOI) system is able to identify objects being manipulated (two red boxes), each of the hands (yellow and green boxes), as well as ignore other objects in the scene (cyan). Using this information we can segment a task into its constituent steps (time-graph below image) for guidance model building or task monitoring.}
 \label{fig:cover}
\end{figure}

Our system (shown in figure \ref{fig:cover}) is not an end-to-end closed box in the fashion of other Deep Learning works. Here, we do use CNNs for the relevant aspects of the task decomposition such as hand and object detection in the wild but combine these with algorithms that help explainability, such as a finite state machine for hand-object interaction control.

Our proposal in this paper detects hand-object interactions from egocentric video of real-world tasks and segments the video into task segments and hand-object events. This is the first crucial step towards the long term objective of assisting in task guidance. Our contributions are threefold: 1. We propose and implement a method that can identify  hands, task-relevant objects and hand-object interactions from egocentric videos in the wild. 2. A method to automatically decompose the input video into its constituent activity steps. 3. Evaluation of our method on non-scripted egocentric datasets quantitatively and qualitatively, including the introduction of a new dataset of Chinese Tea making with expert annotations.

The rest of our paper is organized as follows: After reviewing the related work in section \ref{related_work}, we describe our content authoring paradigm in section \ref{content_authoring}. We then evaluate our results in section \ref{evaluation} and present the conclusions in section \ref{conclusion}.

\section{Related Work}
\label{related_work}
The key contribution of this work is automatic video content authoring with hand-object interaction cues. In this section, we mainly look at work related to the field of unsupervised video content authoring and egocentric hand-object interaction detection.

\subsection{Video Content Authoring and Guidance}
Automating the edit of a video as a tutorial can be traced to systems like Kang and Ikeuchi's \cite{kang1994determination} that classified the hand-object interaction into five phases: approaching, grasp phase, manipulation phase, place object and depart. By analysing fingertip polygon and hand movement speed, breakpoints can be found within an interaction process. Their work is very inspiring, and to our knowledge, it is the first work that utilising hand as a cue to segment an interaction. The work from Mayol and Murray \cite{mayol2005wearable} took one step forward with the objects being interacted with hands extracted by an attention filter that used an in-situ learned skin colour distribution. This allowed to extract a pictorial summary of the interaction. The work from Michihiko et al. \cite{goto2010task} overlays 2D video guidance information in a 3D viewpoint. Overall, there is growing interest in video-based guidance in MR since it can be shown that people can be supported by a system that can display step-wise video guidance on a headset and video guidance reduces the hardware and UX requirements of such system compared to fully-fledged positional 6D AR. The work of Nils and Didier \cite{petersen2012learning}, proposed another unsupervised workflow video segmentation method. Instead of using high-level features like hand and object interactions, they build a distance function with region descriptors to segment the video of a simplified repair task according to the centre crop of each frame. Other systems have looked at how to combine multiple observations for task decomposition. In, Longfei et. al. consider manipulation "hotspots" to identify critical parts of hand-object interactions from multiple observations on a specific object, a sewing machine.
A key first step on all the above systems towards video-based guidance is that the content to be displayed needs to be segmented into coherent steps in advance, something that has been a challenge when considering true non-scripted, multi-object tasks in the wild. On the other hand there have been recent egocentric and non-egocentric deep learning based video segmentation methods such as \cite{li2020ms,wang2020boundary,lea2017temporal, girdhar2021anticipative, wang2020temporal, huang2020improving} that showcase good ability to segment videos into actions. These methods are supervised and heavily rely on the training data, more details can be found in the survey \cite{apostolidis2021video}.

\subsection{Interaction Relevant Object Detection}
Detecting relevant task objects for interaction has received substantial interests in the computer vision community. According to how the interaction is defined, this problem is solved differently. For instance, the early work of \cite{mayol2005wearable} considers as the task-relevant object appearing at the centre of the image frame over a few seconds. While in Teesid et al. \cite{leelasawassuk2017automated}, they hypothesize that the task-relevant object locates within the gaze region that can be predicted with a head-mounted IMU (Inertial Measurement Unit). However, the way of finding task-relevant objects in \cite{leelasawassuk2017automated} is closer to scene recognition. With promising performance achieved by CNN-based object detectors such as \cite{girshick2015fast} \cite{ren2015faster}  \cite{redmon2016you}, it is possible to train an end-to-end model for object-hand interaction detection. Like the work of Dandan et al. \cite{shan2020understanding}, they directly label the hand status and active object for a Faster-RCNN \cite{ren2015faster} based network for detection. This is also what we argue and expand in this work. The interaction/task-relevant object is highly related to the human hand, especially in First Person View. We can assume the hands in view with the right pose relative to the camera mostly belong to the camera wearer. Moreover, the objects interacting with these hands are task-relevant. In our work, we follow a similar paradigm as Shan et al. \cite{shan2020understanding} for task-relevant object detection. But instead of considering all possible hand-object cases, we focus on the first-person view and train the network with our hand object interaction dataset. More details in section \ref{HOI}. 

\begin{figure}[tb]
 \centering % avoid the use of \begin{center}...\end{center} and use \centering instead (more compact)
 \includegraphics[width=0.8\columnwidth]{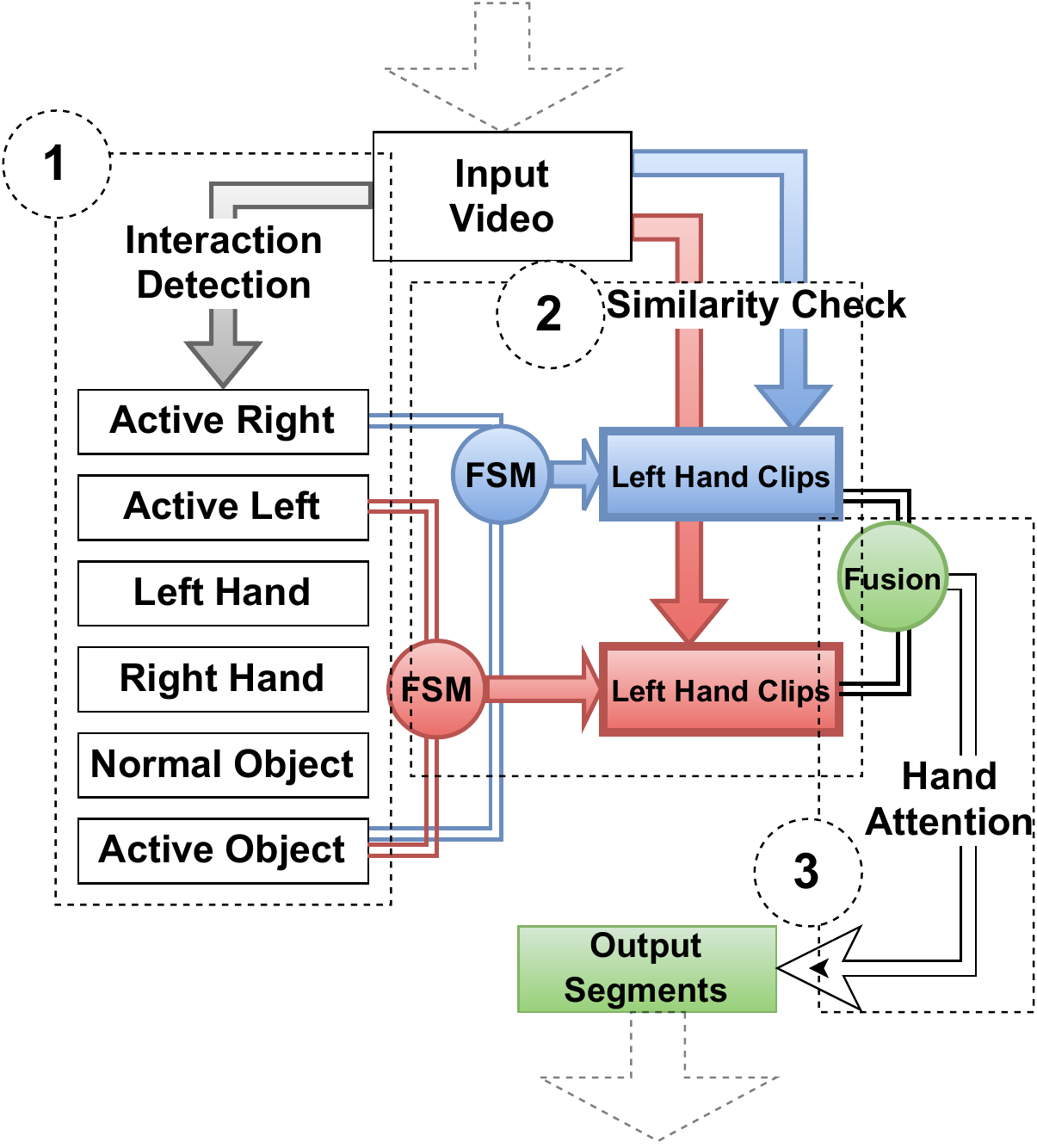}
 \caption{A schematic of our automatic content authoring system. Our system runs in three main stages: (1) Hand-object interaction (HOI) extraction, hand based clips segmentation and segments fusion. With an input video, HOI information is extracted by a Faster-RCNN based detector. (2) By controlling the hand status with FSM (finite state machine) separately, two streams of segmentation can be obtained. (3) Finally, with the hand attention prediction, two streams containing the same step can be fused into one (See Fig \ref{fig:IOU}).} 
 \label{fig:full_system}
\end{figure}

\begin{figure*}[tb]
 \centering % avoid the use of \begin{center}...\end{center} and use \centering instead (more compact)
 \includegraphics[width=2.1\columnwidth]{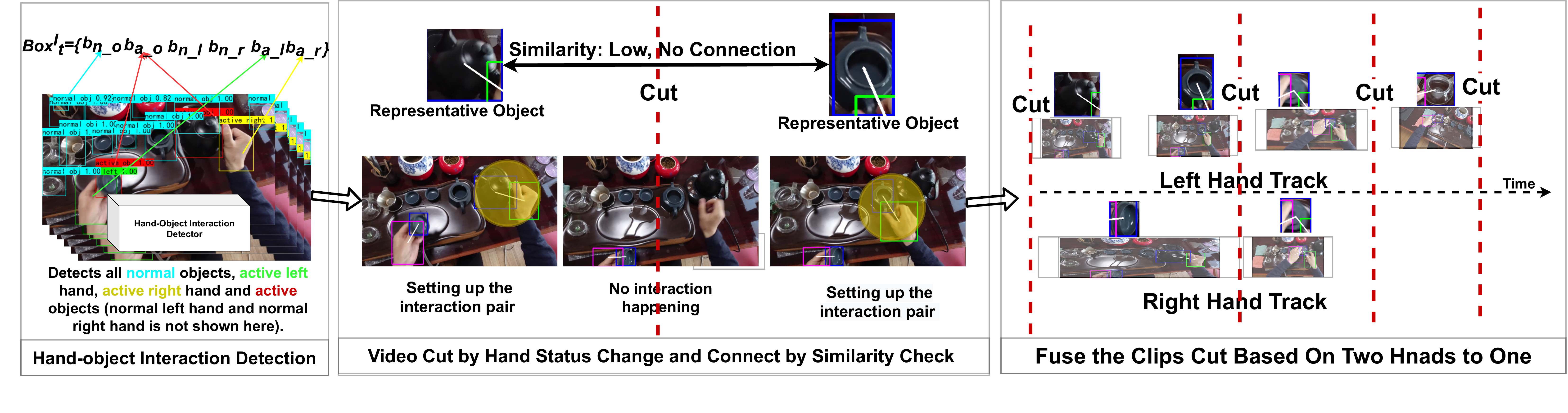}
 \caption{An intuitive illustration of the content authoring process. Left: Hand-Object Interaction information is extracted. The information contains the bounding boxes and classification of all objects and hands. Middle: These extracted information is fed to a finite state machine controlling the hand status (not shown in the picture). Once the hand status changes (e.g., status change from 'interacting with object' to 'idle' or vice versa), we introduce a video segment split. To prevent the over-segmenting of the video, we measure the similarity of two clips based on the automatically obtained object crops and reconnect two adjacent clips if they have high similarity. Right: The video is cut into two streams according to hand and object status. By calculating the IOSA over time, we can unify segments that belong to the same task step. More details on fusion can be found in figure \ref{fig:IOU}} 
 \label{fig:content_authoring}
\end{figure*}

\section{Content Authoring System}
\label{content_authoring}
Figure \ref{fig:full_system} shows a schematic of our system. The workflow of our system is straightforward and runs in three stages: HOI (hand-object information) extraction, hand-based clips segmentation and segments fusion. Given that an input video captured from an expert performs the task in a proper sequence, a Faster-RCNN \cite{ren2015faster} detector is applied to extract the HOI information from each frame. The detected objects and hands are fed into our FSM (Finite State Machine) for hand status control. We can obtain a two-stream initial video segmentation by cutting from the point when the hand status changes from two hands. To prevent over-segmentation of videos, we use a CNN based similarity network to measure the similarity between two adjacent clips. As an outcome, two streams of segmentation (from the left and right hands) are merged into one according to their IOSA (intersect over the smaller area) over time and our hand attention prediction. The result is an input video segmented into its constituent steps.

% In \textbf{guidance delivering} stage, the segmented videos are displayed in sequential order. To help user to identify 'step active object' (object extracted in each tutorial video clip) for each step, the object detector is applied to detect all objects in the scene with similarity check by comparing to crop of 'step active object' in tutorial. The step switch is triggered when newly detected 'active object' is similar to corresponding 'step active object'.  

\subsection{Problem Formulation}
We denote the full input video as $V=\{I_{1},...,I_{T}\}$ which has $T$ frames, for each frame $I_{t}$, we perform our hand-object interaction detector and obtain a set of bounding boxes $Box^{I_{t}} = \{b_{n\_o}, b_{a\_o}, b_{n\_l}, b_{n\_r}, b_{a\_l}, b_{a\_r} \}$ which stand for 'normal object' (an idle object not being interacted with), 'active object' (an object which is being interacted with), 'idle left hand' (left hand which is not interacting with an active object), 'idle right hand', 'active left hand' (object manipulation by left hand), 'active right hand' (object manipulated by right hand). Taking a practical approach to defining hand tasks observed by egocentric video, we assume for a given frame there is {\it at most} $1$ left hand, $1$ right hand and $2$ active objects (i.e. at most one active object per hand). We use a FSM (Finite State Machine) to control the status $S_{l}^{t} $ for a hand. The hand status changes with different input from HOI detection results. The initial segmentation can be obtained by splitting the video where the hand status changes. The initial segmentation of left hand stream can be represented as: $C_{l} = \{c_{1},...,c_{m}\}$, where $m$ is the number of clips. We use a high threshold in HOI detection in each frame to reduce the risk of accumulating the error across the video. However, the video could be over-segmented due to miss detected changes of hand status. As compensation, to reconnect similar clips, we measure the average similarities $\bar{E}$ between the adjacent clips in $C$:
\begin{equation}
    \bar{E}_{m-1,m}=\frac{\sum_{i=1}^{x}\sum_{j=1}^{y}s_{ij}}{i\times j}
    \label{equ:1}
\end{equation}
where $x$ and $y$ are the total number of images chosen from two clips, $s_{ij}$ is a CNN based similarity measure between two images inspired by Siamese \cite{melekhov2016siamese} network. If $\bar{E}_{m-1,m}$ is less than a threshold $T$, we consider the two clips as one. $T$ is found experimentally and based on the performance of all the datasets we use here. The initial segmented clip set of right hand $C_{r}$ can be obtained in the same way. In the end, according to the IOSA (interaction over the smaller area) over time between the two-stream clip sets and the hand attention prediction. We obtain the final segmentation $C_{fused}$.

\subsection{Hand-Object Interaction Detection}
\label{HOI}
\subsubsection{Faster-RCNN Based Detector}
Hand-object interaction is the principal cue in the content authoring of our system. Given one frame from an egocentric view, we consider at most one left hand and one right hand. Each hand can handle at most one object. The hand-object interaction is defined when a hand is physically close to an object. We define a hand that is interacting with an object as an 'active hand'-'active object' pair. Otherwise, the hands and objects in the scene without interaction are defined as 'idle hand' and 'normal object'. We detect every possible 'object' (including hands) in the scene. Each 'object' is classified as 'active left/right hand', 'idle left/right hand', 'active object' or 'normal object'. Our hand-object interaction detector is a trained ResNet-backbone \cite{he2016identity} Faster-RCNN network \cite{ren2015faster}. It takes an image as input and outputs bounding boxes of $6$ possible classes: 'normal object', 'active object', 'active left hand', 'active right hand', 'idle left hand' and 'idle right hand'. As far as we are aware, there is only one public dataset from \cite{shan2020understanding} that has a similar labelling pattern. However, their label is 'hand' centred, the 'active object' is assigned to 'hand', and no 'normal object' label is provided. In addition, most of their data are third-person views which are not suitable for our application. We, therefore, had to do our own labelling for the datasets we use in this paper. We are publicly releasing labels, our new dataset and code accompanying this publication.

\subsubsection{FPV Hand-Object Interaction Dataset}
In total, we label $3K$ images with $6$ labels as per above. Part of the labelled images are chosen from first-person view based datasets, including EPIC-KITCHENS \cite{Damen2018EPICKITCHENS}, GTEA \cite{li2015delving}, and First-Person Hand Action \cite{garcia2017first}. We believe that in some applications, training on a small, specially targeted dataset could make a detector outperforms those trained on a larger general-purpose dataset. In the data selection stage, we intentionally picked images where each:
\begin{itemize}[noitemsep]
    \item [1)]  
    has hands absent or idle
    \item [2)] 
    has one hand interaction and the other idle or absent 
    \item [3)] 
    has two hands interaction separately or together
    \item [4)] 
    has one hands just about to interact
    \item [5)] 
    has hand-object interaction with cluttered background
    \item [6)]
    has hand-object occlusion
\end{itemize}

\begin{figure}[tb]
 \centering % avoid the use of \begin{center}...\end{center} and use \centering instead (more compact)
 \includegraphics[width=1\columnwidth]{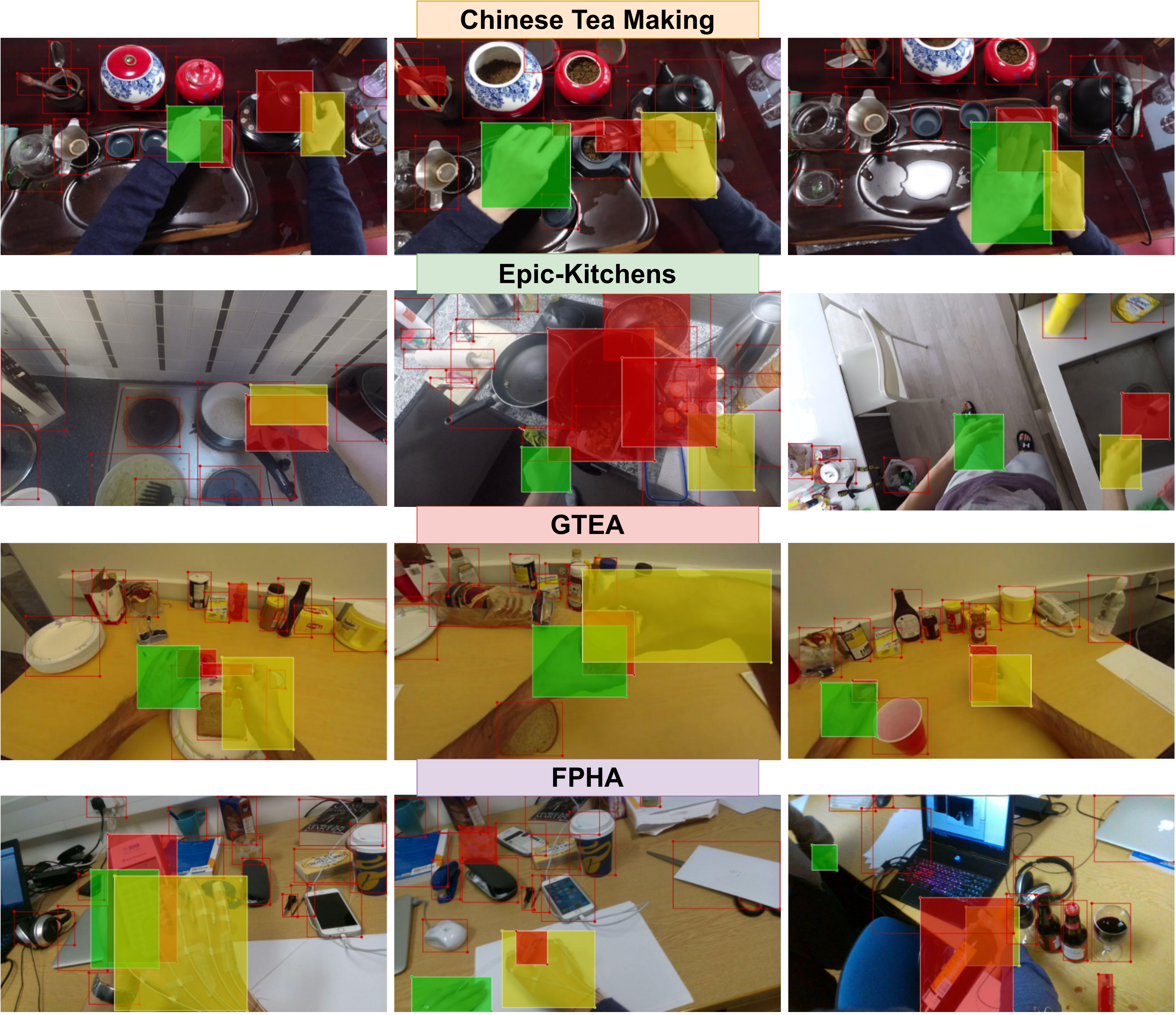}
 \caption{The demo of labelled images from our Chinese Tea Making dataset (row 1), EPIC-Kitchens \cite{Damen2018EPICKITCHENS} dataset (row 2), GTEA \cite{li2015delving} dataset (row 3) and FPHA \cite{garcia2017first} dataset (row 4). Filled green boxes represent left hands, filled yellow boxes represent right hands (activeness is not reflected in the figure). Filled red boxes are 'active object' and other red boxes are 'normal object'.}
 \label{fig:labelling}
 
\end{figure}

Example labelled images are shown in Figure \ref{fig:labelling}. All objects are labelled for each image. We found this very important for fine-tuning the pre-trained RPN (region proposal network) to discover possible objects that have not been seen before. Furthermore, detecting all possible objects is helpful in guidance delivery for step-object compliance checks.

\subsection{Initial Segmentation}
\label{vs}
Here we discuss how we utilize the extracted HOI information for hand status identification.
\subsubsection{Hand Status Control}
\label{sec:hand_status_control}
After obtaining frame-wise detection results $Box^{I_t}$, we build a FSM based hand status model. In the work of Kang and Ikeuchi \cite{kang1994determination}, they consider five phases for hand-object interaction: 'approaching', 'grasp', 'manipulation', 'place down' and 'depart'. This model is object-centred and too idealized to apply in realistic settings. For example, the approach and depart stages are not always clear when the hand is very close to the object without additional information. Also, the boundary between 'grasp' and 'manipulation' is hard to define. Thus, our FSM only considers two different states: ' active' and 'idle' (representing interaction and non-interaction). To have a clear-cut between the two states, we collect and train our hand-object interaction detector with the negative label on 'hand-object approaching' and 'hand-object departing' cases. 

\begin{figure}[tb]
     \centering
     \begin{subfigure}[b]{0.45\textwidth}
         \centering
         \includegraphics[width=\textwidth]{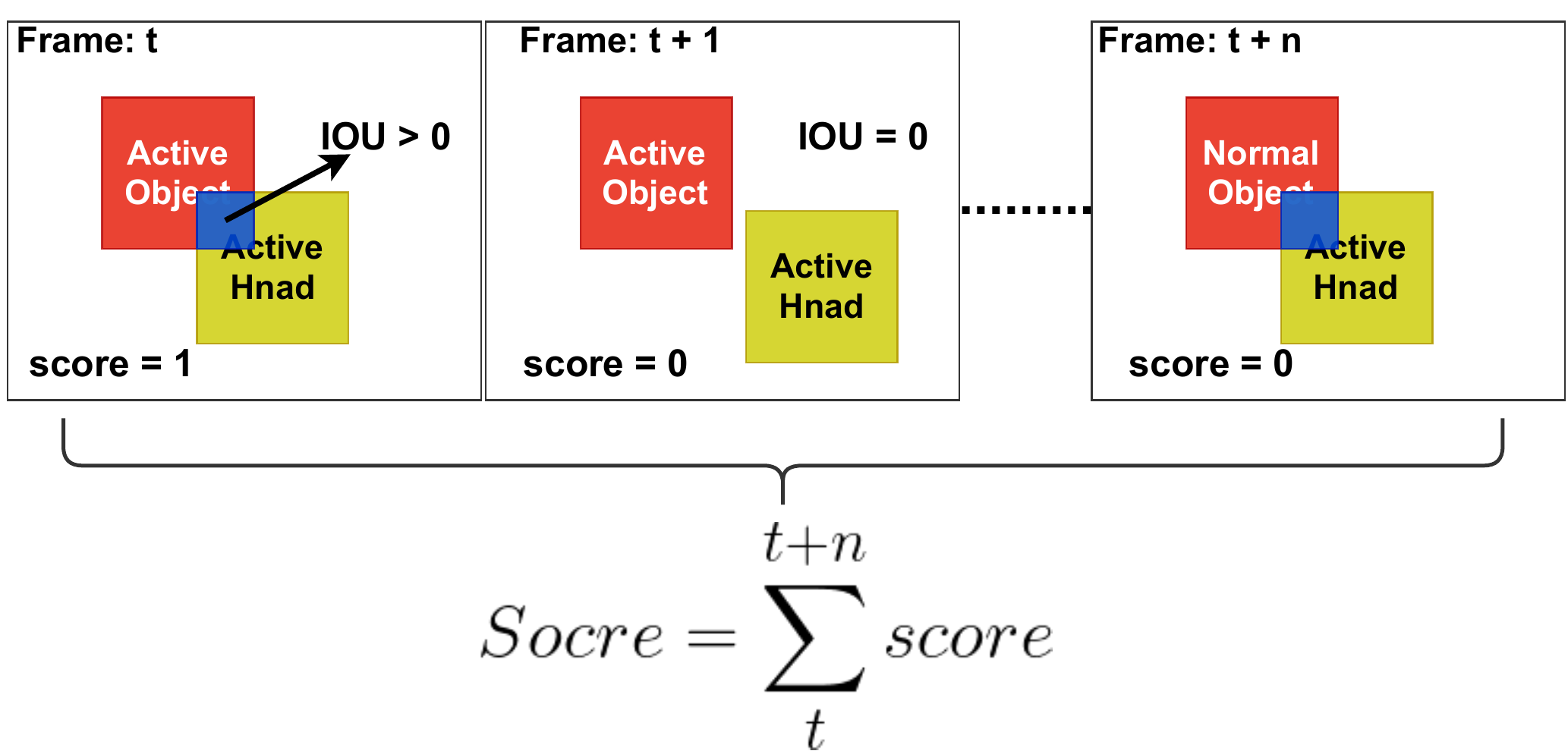}
         \caption{Score calculation}
         \label{fig:sega}
     \end{subfigure}
     \hfill
     \begin{subfigure}[b]{0.5\textwidth}
         \centering
         \includegraphics[width=0.7\textwidth]{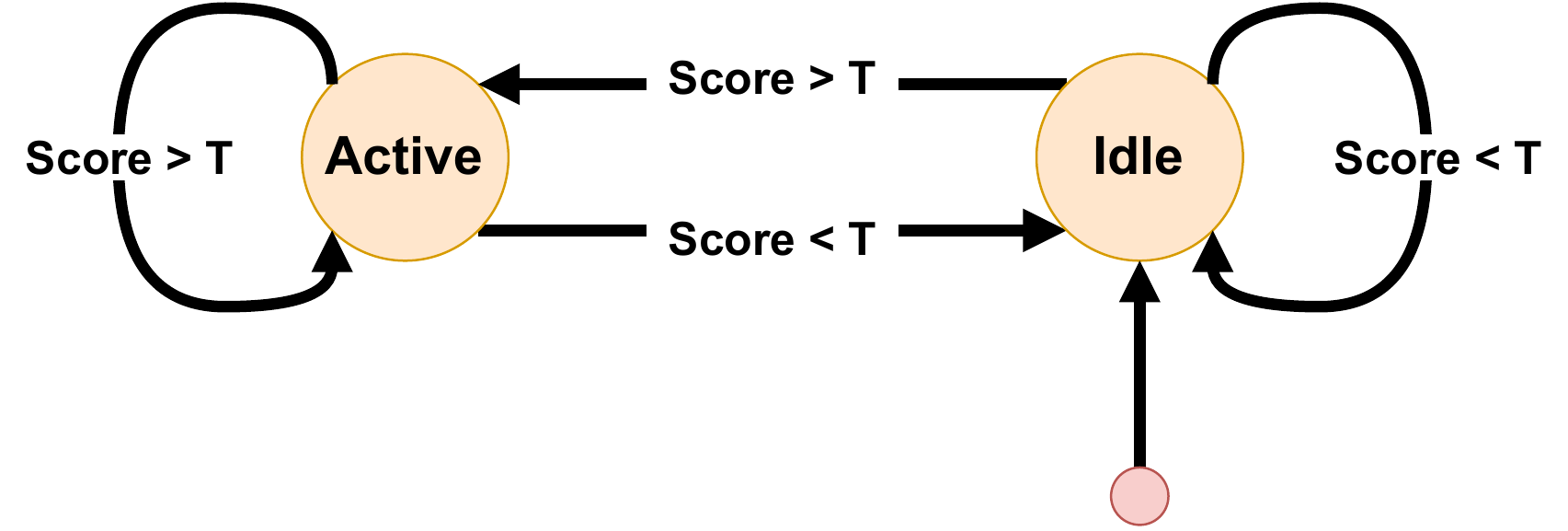}
         \caption{FSM for hand status control}
         \label{fig:segb}
     \end{subfigure}
     \hfill
     \caption{Full hand status determination process}
\end{figure}

\begin{figure}
     \centering
     \begin{subfigure}[b]{0.5\textwidth}
         \centering
         \includegraphics[width=1\textwidth]{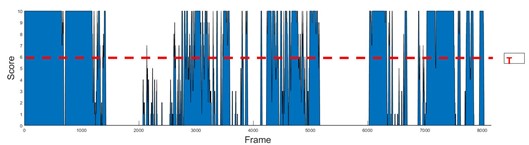}
         \caption{The 'Score' summation within a sliding window across all frames.$T$ is the threshold of hand status change.}
         \label{fig:seg1a}
     \end{subfigure}
     \hfill
     \begin{subfigure}[b]{0.5\textwidth}
         \centering
         \includegraphics[width=1\textwidth]{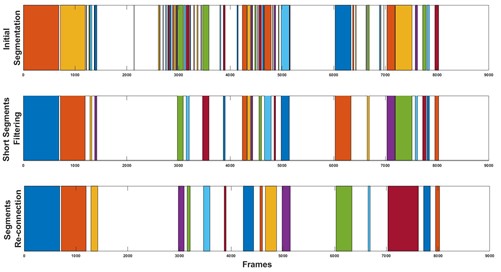}
         \caption{Segmentation results after 'Score' thresholding in (a) (top), the results further optimized by filtering out some segments (middle) and the results after clips re-connection (bottom).}
         \label{fig:seg1b}
     \end{subfigure}
     \hfill
     \caption{The example of video segmentation process. }
\end{figure}

In our FSM, considering the left hand in a frame, if the IOU of bounding boxes between any 'active left hand' and 'active object' is greater than zero, we take the 'score' of this frame as '1' (reflected in Figure \ref{fig:sega}). A sliding window of length $n$ is applied to sum up all 'scores' within it. If the summation from frame $t-n$ to $t$ is greater than a threshold $T$, at the frame $t$, hand status is 'active' (as illustrated in Figure \ref{fig:segb}). Parameters $n$ and $T$ control the sensitivity of hand status change, the smaller they are, the more step clips are obtained. In our implementation, we empirically take $n=\frac{FPS}{6}$ and $T=\frac{FPS}{10}$ (FPS: frames per second of the video and $T \leq n$). Figure \ref{fig:seg1a} shows the summation of scores within the window of length $n$ across a whole video as an example. We take down all the moments the hand status switches from 'idle' to 'active' and from 'active' to 'idle' as 'starts' and 'ends' of segments. In Figure \ref{fig:seg1a}, the status change happens at the moment the score summation equals $T$.

Having obtained an initial segmentation $C_l^{initial}$ which is shown on the first subplot of figure \ref{fig:seg1b}, we can see that many segments only last for a relatively short period. This can be caused by the uncertainty when status change and usually happens at the beginning and end of a step. We regard the segments which are less than half a second as a false positive segmentation and remove them directly. The result after short segment filtering $C_l^{filtering}$ is shown on the second subplot of Figure \ref{fig:seg1b}.

\subsubsection{Clips Re-connection by Similarity Check}
Compared with the number of frames per second, the sliding window length and threshold chosen to determine the hand status is relatively small. This makes the hand status prediction more sensitive to ballistic hand movements. At the same time, more incorrect step splits could be introduced. We thus follow a 'break and reconnect' strategy to solve over-segmentation. In other words, we check the similarity of the 'active objects' extracted from two adjacent clips and determined whether to link them up. The clips that have links between them are reconnected as one segment.

Recall that the process of determining a hand-object interaction pair in section \ref{sec:hand_status_control}. If a frame is considered as having an interaction on a given hand, we assume there is only one active object for that hand. For each hand, every frame in $C_l^{filtering}$ has a corresponding object crop. As per \ref{equ:1}, the similarity measure averaged image similarities between two adjacent segments by densely calculating the crop-wise similarity. Considering the appearance of the object's crop within the segment is not always the same due to, e.g. pose changes, the similarity measure only happens between the $20\%$ of the ending frames for the foregoing segment and the $20\%$ beginning frames for the posterior segment.

\begin{figure}[tb]
 \centering % avoid the use of \begin{center}...\end{center} and use \centering instead (more compact)
 \includegraphics[width=1\columnwidth]{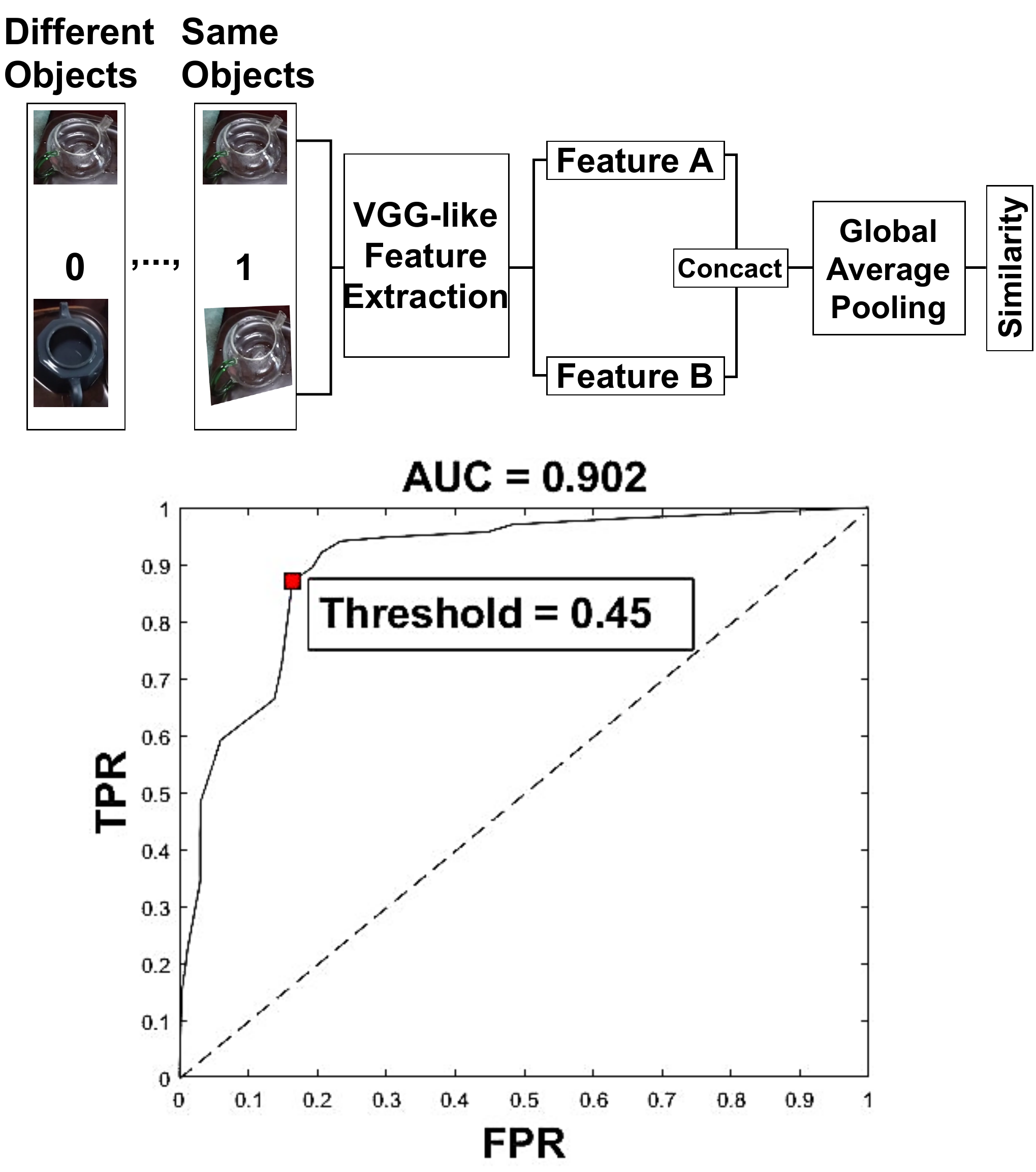}
 \caption{The Siamese net based image wise similarity measure network (top). And the threshold selection for similarity. (bottom)}
 \label{fig:similarity}
\end{figure}

The image-wise similarity is achieved by training a Siamese network \cite{melekhov2016siamese}. As shown in Figure \ref{fig:similarity}, two images pass through the same feature extractor. For positive samples, we choose an image crop. The training pair is generated by randomly applying image transformation warps and flips. The negative samples come from different object crops. We use \textit{Global Average Pooling} \cite{lin2013network} to get the result and \textit{Binary Cross-Entropy Loss} as loss function. Since our HOI dataset has over $20K$ object instances, it can be used to train our similarity check network.

We use ROC (Receiver operating characteristic) curve analysis to find the best threshold for segment re-connection. We manually select $5$ objects and find $10$ instances for each of them from different frames in our HOI dataset. For each object class, we run $1000$ times similarity check (the positive ground truth rate is set to $50\%$) by selecting images pairs randomly. With different thresholds from $0-1$, we plot the ROC curve shown on the bottom of Figure \ref{fig:similarity}.

\begin{figure}[tb]
 \centering % avoid the use of \begin{center}...\end{center} and use \centering instead (more compact)
 \includegraphics[width=1\columnwidth]{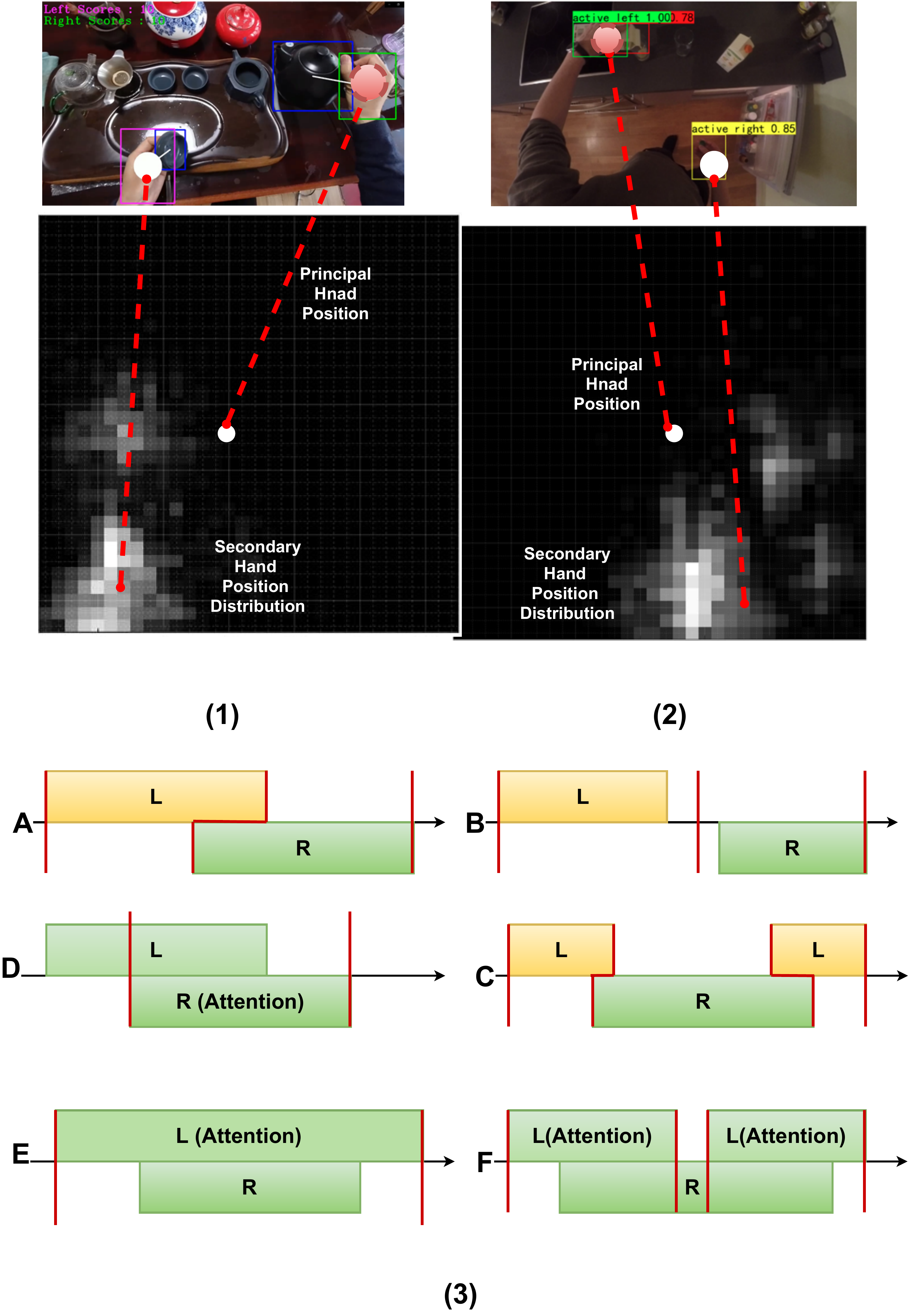}
 \caption{(1) Shows the left-hand position distribution when the right hand is considered as principal hand (the hand dominates the action). (2) shows the right-hand position distribution when the left hand is considered as the principal hand. We take the centre of the hand bounding box as hand position. The position of the right/left hand on (1)/(2) is normalized to the centre of the hand attention map (white dot), and all distances are normalized within $32\times32$ (the map size). (3) shows the cases we consider in segments fusion: A, B and C are the cases the IOSA (intersect over the smaller area) of two segments is less than $0.5$. D, E and F are the cases IOSA is greater than $0.5$, the hand with attention dominates the segmentation.}
 \label{fig:IOU}
\end{figure}
\subsection{Fusion of Two Hand Streams}
 To get the final task step decomposition segmentation, the results from left and right hands need to be combined into one. We hypothesize that for a single hand-object interaction process, the attention can only be put on one 'active object'-'active hand' pair (there are at most two processes in one frame). We define the attention as 'hand attention', and the hand gained attention as 'principal hand', correspondingly, the hand with less or no attention is called 'secondary hand'.  
 %From our life experience and observations from existing datasets. Like people can't play piano with both hands without training, We found it is hard for people to interact (not just hold) wilt different object simultaneously. We hypothesize that there is at most one object has attention in single frame. This a sub-classification of the case that two hands status are both interacting with different objects. 

\begin{figure}[tb]
 \centering % avoid the use of \begin{center}...\end{center} and use \centering instead (more compact)
 \includegraphics[width=1\columnwidth]{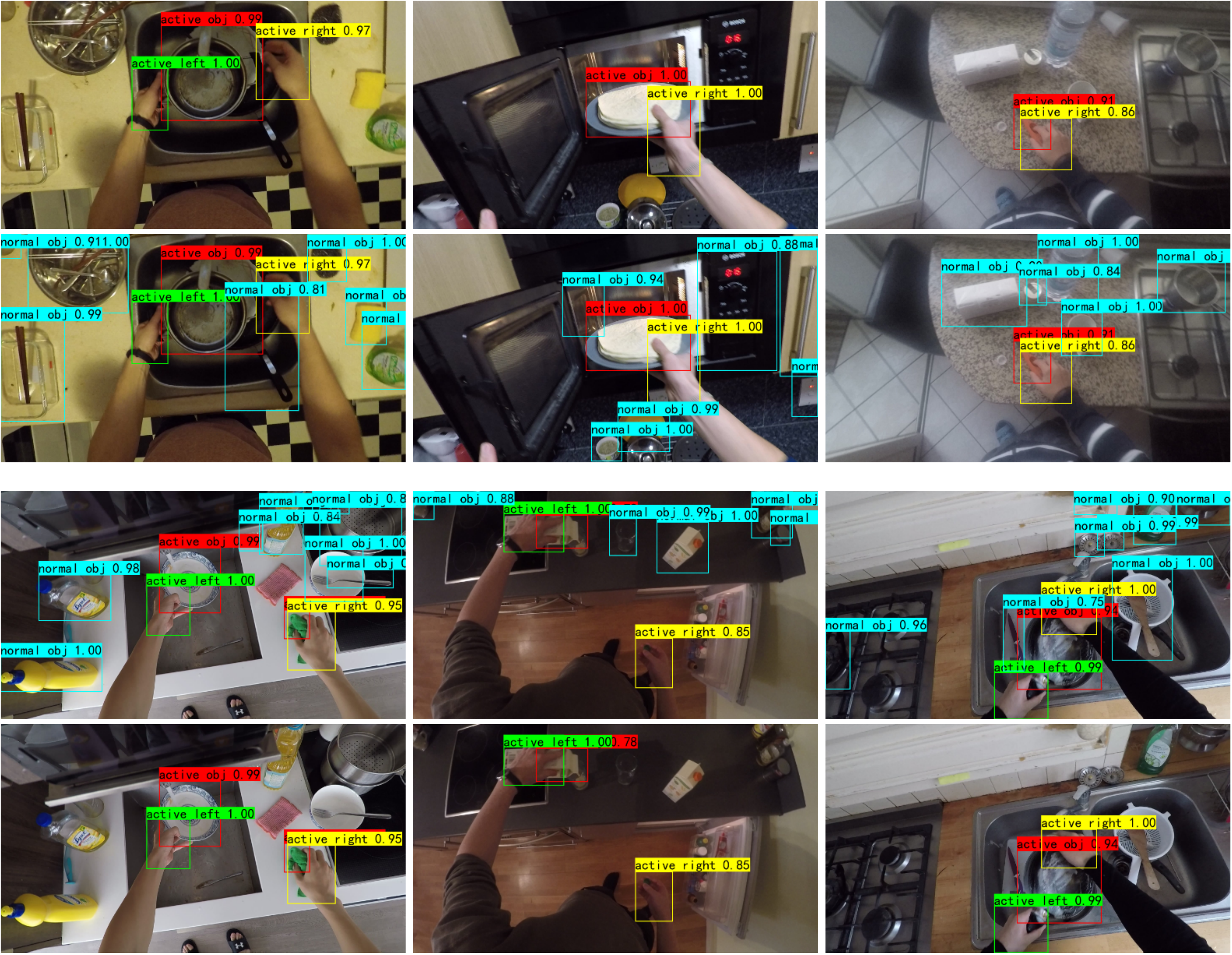}
 \caption{Some examples of hand-object interaction results. Top rows has the results including 'normal object' detection.}
 \label{fig:hoi-rst}
\end{figure}

We hypothesise that the hand attention as defined here is highly related to the relative position of the two hands, especially when both of them are interacting with different objects. To learn the distribution characteristics of the hand-based attention mechanism, we follow the rules of determining interaction status to extract all the cases in our FPV-HOI Detection dataset when both hands are in 'interaction with objects'. Then the attention on hand is labelled manually (labelling the hand that dominates the action). We plot the relative position of two hands when they are all recognized as in 'interaction status'. For example, ($1$) in figure \ref{fig:IOU} shows the left-hand position distribution when the right hand is normalized to the centre of the hand attention map. From distributions ($1$) and ($2$) in figure \ref{fig:IOU}, we observe that the secondary hand in most of the cases is located in a position that is spatially lower to that of the principal hand. While we use multiple and varied hand activities in our training datasets (GTEA\cite{li2015delving}, Epic-Kitchen \cite{Damen2018EPICKITCHENS}, FPHA \cite{garcia2017first} and our Chinese Tea Making datasets). We agree that there is the scope that more data labelled from multiple persons will inform better the distribution of these events.

%This conclusion is determined by single-person labeling on our dataset. To generalize it, more labels from different people and more experiments are required. In this work, we simply use the conclusion and evaluate it together with video segmentation

\begin{figure*}[tb]
 \centering % avoid the use of \begin{center}...\end{center} and use \centering instead (more compact)
 \includegraphics[width=2\columnwidth]{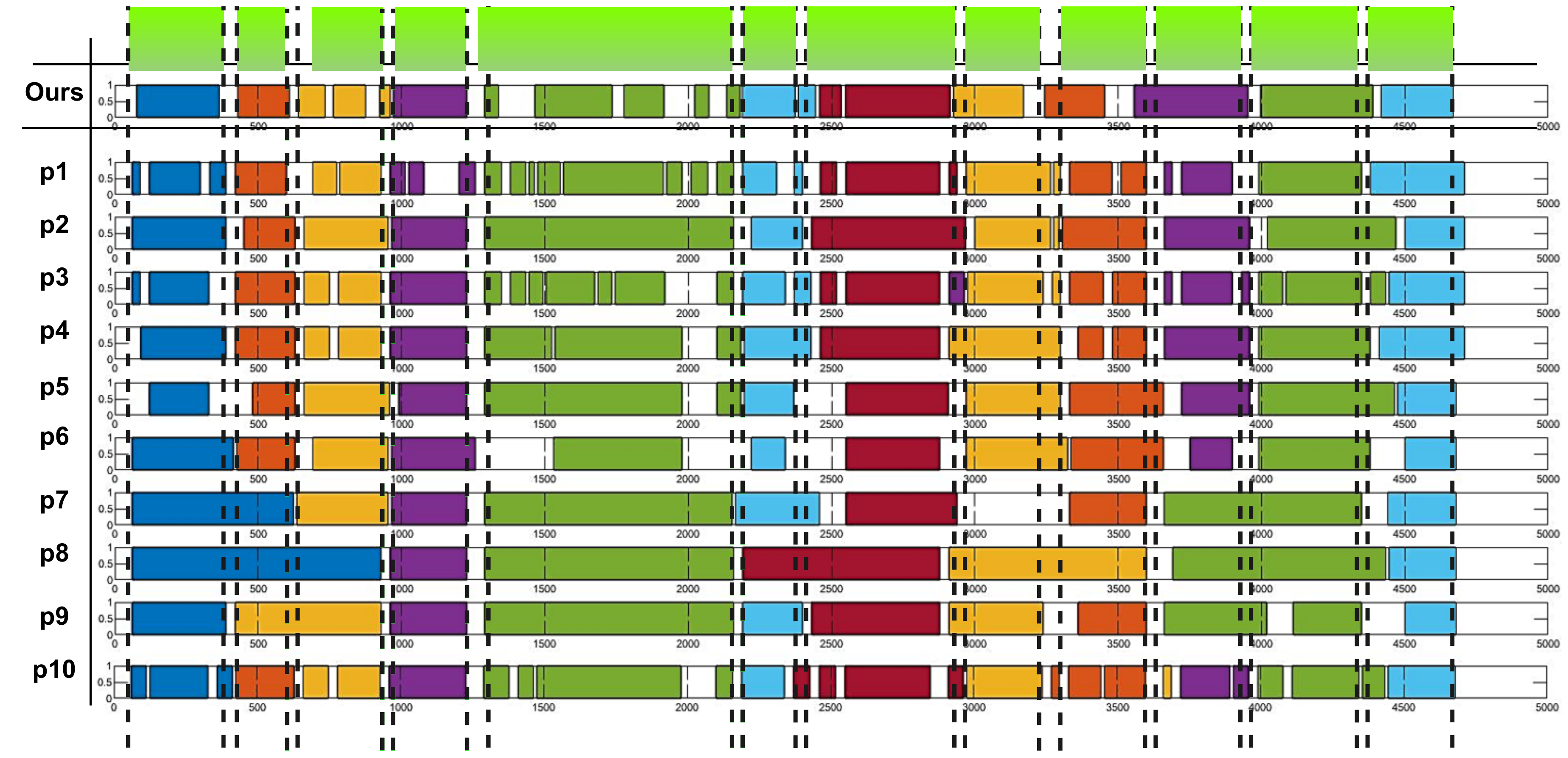}
 \caption{An example segmentation result on a Chinese Tea Making video. It shows the results from our paradigm and the annotations from $10$ invited participants. The numbers in top row represent the instruction based expert's segmentation.}
 \label{fig:rst}
\end{figure*}

% \subsection{Guidance Delivering}
\section{Evaluation}
\label{evaluation}
In this section, we use different experiments to (1) evaluate the performance of our hand-object interaction detector quantitatively. (2) quantitatively evaluate the performance of our video segmentation results on GTEA \cite{li2015delving}. (3) evaluate our Chinese-tea making based on expert instruction segmentation and participants' annotations. 

\subsection{Dataset Selection}
Here we list all the datasets we use for training and evaluation.
\begin{itemize}
    \item \textbf{FPV-HOI}. This is the dataset we collected $3000$ images from EPIC-KITCHENS \cite{Damen2018EPICKITCHENS}, GTEA \cite{li2015delving} and FPHA \cite{garcia2017first} and labelled by us. It contains $6$ kinds of labels. We train our HOI detector with this dataset.

    \item \textbf{EPIC-KITCHEN 'unseen'}. EPIC-KITCHEN is an egocentric dataset on kitchen work recording. Action labels are provided, but there are no scripted tasks included which prevent its use in our study here. However we use the images in its 'unseen' kitchen as the test set for our HOI detector.

    \item \textbf{GTEA}. GTEA is a dataset that has $28$ egocentric videos of sequential-step task performing. The action annotations are given in a verb-noun form. We use it for our video segmentation evaluation.
    
    \item \textbf{Chinese Tea Making}. Chinese Tea Making Dataset is an egocentric dataset collected by us. Three experts are invited to perform Chinese tea making twice without any prior knowledge of computer vision. Each video lasts for about $2.5$ minutes containing $12$ different steps. What makes this dataset complex is the cross hand-object manipulation for each step. Multiple objects and hands may have interaction.
\end{itemize}

\subsection{Evaluation of Hand-Object Interaction Detection}
\label{evaluate HOI}
There are several hand-object interaction datasets of interest for our work. Such as EPIC-KITCHENS \cite{Damen2018EPICKITCHENS}, VLOG \cite{fouhey2018lifestyle}, GTEA \cite{li2015delving}, 100DOH \cite{shan2020understanding} and AVA \cite{gu2018ava}. The EPIC-KITCHEN provides auto-labelled object bounding boxes only. In the case of 100DOH, it provides labelling of full-hand-state and interaction objects, but most of the data is third-person based and cannot be distinguished from the available labels. Although our hand-object interaction detector has qualitatively good performance on different datasets (Fig \ref{fig:hoi-rst}), building a general HOI detector is not the scope of this work. To make sure there are enough hand-object interactions in our testing set, we select $300$ images from the 'unseen' set (the set not in our training set) of EPIC-KITCHENS and only label with hand status and active objects (compared with training images, no 'normal object' is labelled in a testing set). The results are shown in table \ref{tab:hoi}.        

\begin{table}
\centering
\caption{The results of HOI detector on the 'unseen' subset from EPIC-KITCHENS \cite{Damen2018EPICKITCHENS}. In the first row, AH: 'active hand', AO: 'active object' 'HOI': The hand object interaction. The leading number means the number of instances in total. From the second row, TP: number of true positives, FP: number of false positives, TPR: True positive rate. Our HOI detector achieves $97.32\%$ precision which means the possibility a frame is wrongly classified as HOI is low.}
\begin{tabular}{ |l|c|c|c|c| } 

\hline
     & 385 AH & 322 AO & 385 HOI \\
\hline
% \multirow
TP        & 341 & 263 & 291\\ 
FP        & 45 & 31 & 8\\ 
TPR       & 88.57\% & 81.63\% &75.58\%\\ 
Precision & 88.34\% & 89.46\% & \textbf{97.32\%}\\
\hline
\end{tabular}
\label{tab:hoi}
\end{table}

On the testing data, we achieve $97.32\%$ precision (true positive over true positive plus false positive) on hand-object interaction. We use a relatively high threshold for the detector because the accuracy loss caused by false negatives for a single frame can be compensated by using video temporal information via hand status control. The figure \label{fig:hoi-rst} shows example results of our HOI detector. We detect all possible objects and hand-object interactions.

\subsection{Evaluation on Video Segmentation }
We mainly evaluate our task segmentation workflow on GTEA \cite{li2015delving} and our Chinese Tea Making dataset since, in contrast to EPIC-KITCHENS, they do have repeated videos for the sequential-step activities. GTEA is a breakfast-making dataset that provides two kinds of labels: action-based and object-based. The action-based label is composed of a verb and a noun. For example, 'take cheese', 'open cheese' and 'put cheese on the bread' are different segments in the action-based label. While in the object-based label, they are all 'cheese' relevant actions and belonging to the same segment. The object-based segmentation has fewer breaks across the video, which is more natural for a human to understand, and we follow this labelling style in GTEA dataset evaluation. Chinese Tea Making dataset is more realistic and thus more complex than GTEA. It is hard to define the ground truth of video segmentation based on either 'active objects' or actions because two hands may have different in-hand objects and actions on a timestamp. We use the instruction based annotation (segmented by an expert) and participants' annotation as two distinct references to evaluate our results on the Chinese Tea Making dataset. 

We use a segmentation $\textbf{F1}$ score with overlapping thresholds (IOU) at $10\%$, $20\%$ and $50\%$ as quantitative evaluation which is proposed by \cite{lea2017temporal}. As for processing time, our video segmentation currently runs in an offline manner. With an 'i7-6700' CPU laptop and 'Quadro M2000' GPU. The processing time on hand-object detector is about $2$ frames per second. We expect this can be optimized in a number of ways.

\subsubsection{Results on GTEA dataset}
As mentioned above, GTEA is a dataset that has action and in-hand object labels with time segments. Because our method is in-hand object-centred, we use the object-based label as ground truth. This level of annotation is also the reason why the comparisons with other works are not performed. We report $F1$ score in table \ref{tab:gtea_f}. 
\begin{table}
\centering
\caption{The $F1$ score with overlapping thresholds (IOU) $10\%$, $20\%$ and $50\%$ on GTEA dataset.}
\begin{tabular}{ |l|c|c|c| } 

\hline
     GTEA Results & F1@$10\%$ & F1@$30\%$ & F1@$50\%$ \\
\hline
% \multirow
     & 81.67\% & 73.01\% &69.50\%\\ 
\hline
\end{tabular}
\label{tab:gtea_f}
\end{table}

\begin{figure}[tb]
 \centering % avoid the use of \begin{center}...\end{center} and use \centering instead (more compact)
 \includegraphics[width=1\columnwidth]{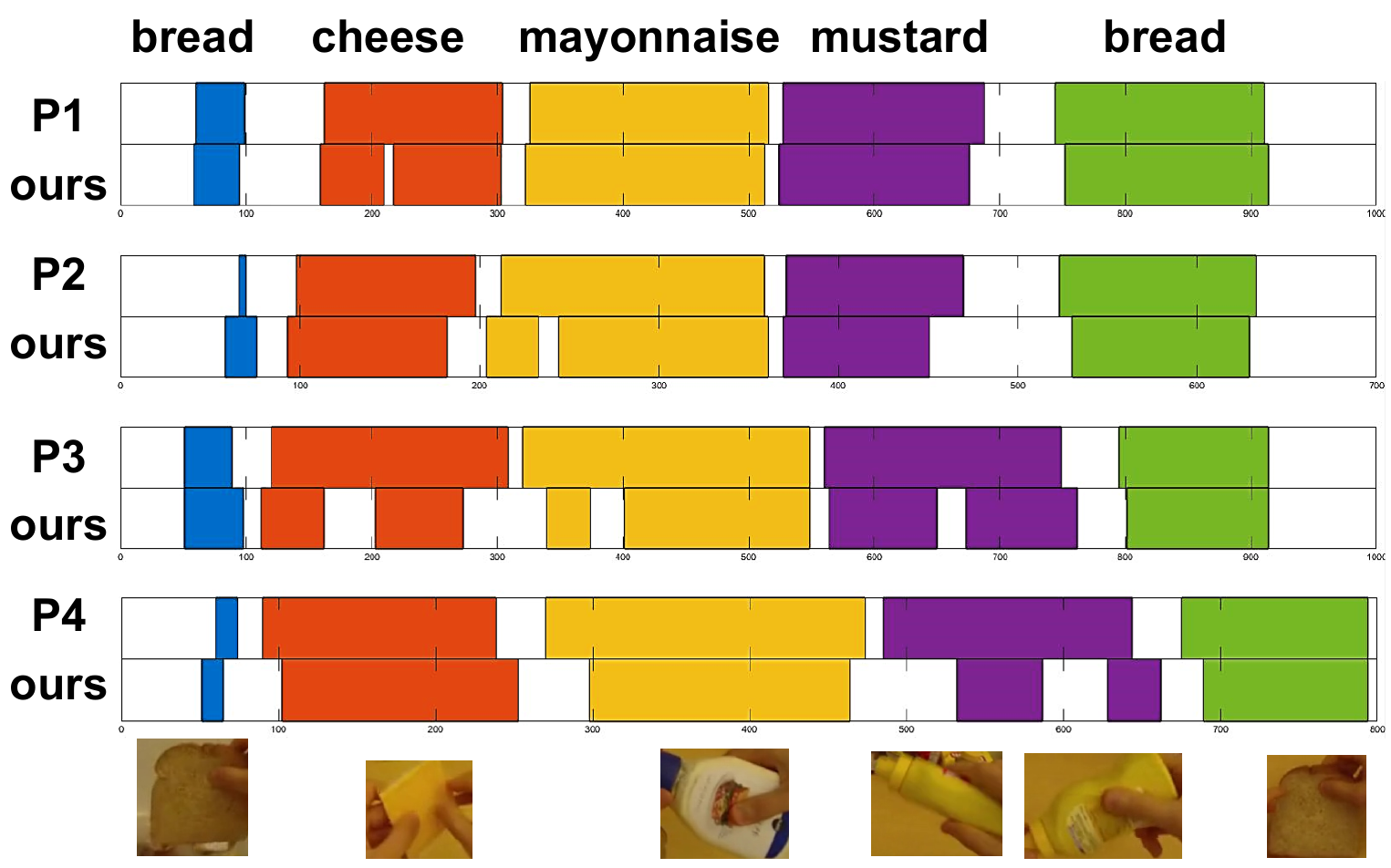}
 \caption{Example result of 'cheese sandwich making' from GTEA.}
 \label{fig:gtea_rst}
\end{figure}

Figure \ref{fig:gtea_rst} depicts a typical result of 'cheese sandwich making' in GTEA. Our prediction has accurate separation of different in-hand object manipulation and achieves $81.67\%$, $73.01\%$, $69.50\%$ $F1$ score, respectively. The corresponding image crops shown in the bottom of \ref{fig:gtea_rst} come from the object crop of the middle frame in each segment and align the nouns in the ground truth. We trace the reason for having an extra segment in cheese preparation and find it is caused by the failure detection of cheese and the similarity check between the two segments. For the highly deformable object like cheese, the 'cheese' and the 'In-hand cheese' could be two different objects from the perspective of the detector. This could be solved in future work by modelling the difference of an object in different states.

\subsubsection{Results on Chinese Tea Making Dataset}
 To have a clearer understanding of this dataset, we list all the objects involved and steps for performing the dataset in table \ref{tab:steps}. 

\begin{figure}
 \centering % avoid the use of \begin{center}...\end{center} and use \centering instead (more compact)
 \includegraphics[width=1\columnwidth]{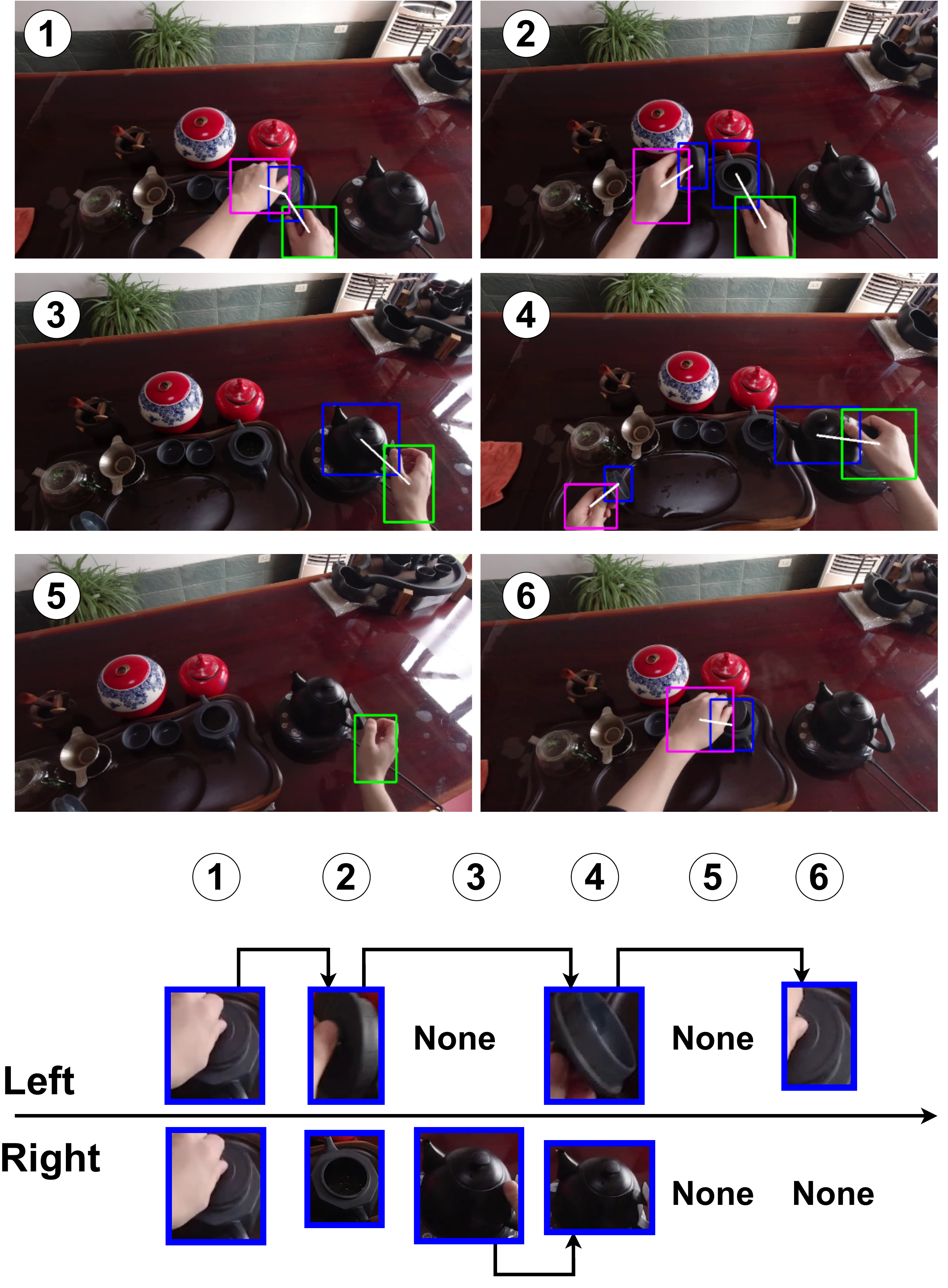}
 \caption{The visualization of step 'pouring water form the kettle to the teapot'.}
 \label{fig:CTM}
\end{figure}

The steps shown in the table \ref{tab:steps} are written and agreed upon by all experts, which summarizes the process of Chinese tea making. However, it is neither action by action nor an object by object instruction. For example, in the Chinese tea making process, the step $1$: 'pour the tea from the kettle to the teapot' is not a step with a single action or a single object. As shown in figure \ref{fig:CTM}, our system roughly split the first step: 'pour the tea from the kettle to the teapot' into $6$ stages according to the clips from left and right hand before fusion (the arrow show the clips to be connected). 'Teapot lid', 'teapot', 'kettle' and both hands are involved in this single step. Coincidentally, the left hand's object (teapot lid) does not change through this step. By conducting crop similarity check, all clips from the left hand are reconnected into one. By the rules of segments fusion in section \ref{vs} and the case $E$ in figure \ref{fig:IOU}. The long segment dominants in video segmentation. However, in another video performed by another expert, the step $1$ is implemented by a single right hand, leading to different segmentation results. It is subjective which segmentation approach is better in terms of guidance without further human involved experiments. Thus, the instruction is used as a reference instead of definite ground truth.

% The scoop should be taken out from the pot in advance and it is not reflected in instruction. %Moreover, it doesn't have details of how a step is tackled with both hands. Thus, the instruction is used as a reference in our results evaluation instead of a proper ground truth.

\begin{table}
  \centering
\caption{Instruction steps in Chinese Tea Making dataset}

\begin{tabular}{ |l|c|l|c|} 

\hline
\textbf{Object} & \textbf{Number} &\textbf{Object} & \textbf{Number} \\
\hline
% \multirow
teapot        & 1 &  jar & 7\\ 
teapot lid       & 2 & jar lid & 8 \\ 
kettle       & 3 & wooden spoon & 9\\ 
teacup       & 4 & tea filter  & 10\\ 
brush       & 5 & glass teacup & 11\\ 
metal tea scoop       & 6 & glass teacup lid   & 12\\ 
% jar                 & 7 \\ 
% jar lid             & 8 \\ 
% wooden spoon       & 9 \\ 
% tea filter       & 10 \\ 
% glass teacup       & 11 \\ 
% glass teacup lid   & 12 \\ 
\hline
\end{tabular}
\begin{flushleft}
The steps written with object number and example hand-object interaction case (the hand usage differs in different video):
\end{flushleft}
\begin{tabular}{ |l|c|c|c|} 

\hline
\textbf{Steps} & \textbf{Left Hand} & \textbf{Right Hand}  \\
\hline
% \multirow
1.pouring water form 3 to 1        & 2 & 3 \\ 
2.pouring water from 1 into 4        & $\backslash$ & 1 \\ 
3.pouring water out of 4        & $\backslash$ & 4 \\ 
4.brushing the table            & $\backslash$ & 5 \\
5.scoop the tea from 7 to 1            & 8, 6 & 9, 2 \\
6.pouring water form 3 to 1        & $\backslash$ & 3, 2 \\ 
7.pouring tea from 1 into 11    & 12 & 10, 1, 11 \\ 
8.pouring tea from 11 into 4    & 11 & 11 \\ 
9.pouring tea out of 4          & $\backslash$ & 4 \\
10.pouring water form 3 to 1        & $\backslash$ & 3 \\ 
11.pouring tea from 1 into 11    & 2, 12, 10 & 1 \\ 
12.pouring tea from 11 into 4    & 11 & 11 \\ 

\hline
\end{tabular}
\label{tab:steps}
\end{table}

In order to have a more objective view of our results, we invite $10$ participants to segment into steps the $6$ videos in the Chinese Tea Making dataset. They are required to split the videos into different steps. Each step can be expressed with a 'verb' plus 'noun' pair. Gaps are allowed in segmentation. Figure \ref{fig:rst} shows our segmentation result and participant's annotations for one video. The numbers from $1-12$ represent the instruction from the expert's segmentation, and $p1-p10$ are the results from the participants. As expected, they show some consensus and some disagreements with expert's segmentation. Like step $3$, the actual content is pouring out water from two teacups. It is reasonable to either consider this step as a segment or split it into two. In table \ref{tab:Chinese}, we report the $F1$ score of our results on expert's segmentation, average results on participants' annotations and the average results between participants' and expert's. The score shows our result has close parallels with both expert's annotation and participants' annotations.

\begin{table}
\centering
\caption{The $F1$ score with overlapping thresholds (IOU) $10\%$, $20\%$ and $50\%$ based on instruction and human labelled references.}
\begin{tabular}{ |l|c|c|c|c| } 

\hline
     Chinese Tea Results& F1@$10\%$ & F1@$30\%$ & F1@$50\%$ \\
\hline
% \multirow
ours on instruction        & 75.52\% & 70.74\% & 66.28\%\\ 
ours on $10$ labels       & 72.34\% & 68.30\% & 63.95\%\\ 
$10$ labels on instruction       & 69.19\% & 61.76\%  & 54.55\%\\ 
\hline
\end{tabular}
\label{tab:Chinese}
\end{table}

Our initial motivation for this work was to close the loop and provide video guidance for people to achieve tasks, such as Chinese Tea making.
%To further evaluate the segmentation performance from a perspective of guidance. People supposed to be asked to perform Chinese tea making with the segmented video clips. 
However, due to the COVID-19 restrictions, we cannot conduct experiments with volunteers. Instead, we opted for show a segmentation result with the most segments and invite a new participant to transcribe the content of each video clip with verbs and nouns. If they think the clip is not clear enough or cannot express the segment with a verb-noun pair, the clip is labelled as \textbf{\textit{not clear}}. This approach offers an indication of the quality of the potential guidance and task decomposition.

\begin{table}

\begin{tabular}{ |l|l|} 
\hline
\multicolumn{2}{|c|}{\textbf{\textit{Human description based on our results}}} \\
\hline
% \multirow
1 &'Fill the teapot with hot water' \\
\hline
2 &'pour the hot water into teacup'  \\
\hline
3 &'pour out the water from teacup' \\
\hline
4 &'\textit{\textbf{not clear}}', 'take the brush', '\textit{\textbf{dip the brush}}' \\
%  &'take the brush' \\
%  &'\textit{\textbf{dip the brush}}' \\
\hline
5 &'take the lid of the jar', '\textit{\textbf{not clear}}', 'take the spoon' \\
%  &'\textit{\textbf{not clear}}' \\
%  &'take the spoon' \\
 &'\textit{\textbf{not clear}}', 'scoop the tea from the jar' \\
%  &'scoop the tea from the jar' \\
%  &'remove the lid of the teapot' \\
 &'remove the lid of the teapot', 'put the tea into teapot' \\
 &'\textit{\textbf{not clear}}', 'put the lid back to the jar' \\
%  & \\
\hline
6 &'pour water to the teapot' \\
\hline
7 &'put the lid back to the teapot'\\
 &'put the filter onto the glass teapot' \\
 &'filtering the tea' \\
\hline
8 &'pour the tea into the teacup' \\
\hline
9 &'pour out the tea from teacup' \\
\hline
10 &'fill the teapot with hot water' \\
\hline
11 &'filtering the tea' \\
\hline
12 &'put back the lid of glass teapot' \\
 &'pour the tea into the teacup' \\
\hline
\end{tabular}
\caption{The description of a participant after watching the segmented video clips. The numbers are the instruction based segmentation which are used as a reference.}
\label{tab:scribbed_steps}
\end{table}

From the table \ref{tab:scribbed_steps}, except for the step $4$ and $5$, descriptions align well to the instructions. In step $4$, the actual step is ‘brushing the table’ (in order to get rid of the water poured out from the previous step). Due to the false negatives on 'brush' detection, the interaction status is not recognized correctly. While the step $5$ is over segmented. Because there are many object transfers between both hands exist in this step, some clips don't show clear intentions.     

\section{Conclusion}
\label{conclusion}
Task guidance is an essential application for MR systems. In this paper, we focus on the critical competence of task decomposition to support video-based guidance. 
Our approach uses CNN modules and explainable Finite State Machines to extract the steps that decompose hand activities automatically. 
We evaluate in real tasks on both public and specifically collected video datasets. Our approach combines hand detection and objects similarity check module to edit videos into steps automatically. We address egocentric step segmentation and segment fusion problems by analysing hand-object interactions and reasoning about hand and object activity within frames. Our results show that we can achieve high precision in step decomposition for unseen tasks, and our results agree within the levels of subjectivity that volunteers can judge.
We believe MR systems need to tackle the crucial problem of automated content editing, and methods that develop this aspect for real-life activities will help get us closer to the broader adoption of MR systems. In future work, we will explore methods of delivering the extracted video content to users for task guidance.

\bibliographystyle{abbrv-doi}

\bibliography{template}
\end{document}